\documentclass[conference,compsoc]{IEEEtran}
\usepackage{amsfonts}
\usepackage[margin=1.2in]{geometry}
\usepackage{graphicx}
\usepackage{multirow}

\usepackage{lastpage}
\usepackage{fancyhdr}
\ifCLASSOPTIONcompsoc
  \usepackage[nocompress]{cite}
\else
  \usepackage{cite}
\fi
\ifCLASSINFOpdf
\else
\fi
\usepackage{colortbl}
\usepackage{color}
\usepackage{array, makecell}
\usepackage{boldline}
\usepackage[x11names, table]{xcolor}

\newcolumntype{?}[1]{!{\vrule width #1}}
\definecolor{colT1}{rgb}{0.95,0.88,0.2}
\definecolor{colT2}{rgb}{0.98,0.95,0.8}
\definecolor{colT3}{rgb}{0.0,0.0,0.0}

\begin{document}
%
\title{Vehicle Shape and Color Classification Using Convolutional Neural Network}

\author{\IEEEauthorblockN{Mohamed Nafzi}
\IEEEauthorblockA{Facial \& Video Analytics\\
IDEMIA Identity \& Security Germany AG\\
       mohamed.nafzi@idemia.com}
\and
\IEEEauthorblockN{Michael Brauckmann}
\IEEEauthorblockA{Facial \& Video Analytics\\
IDEMIA Identity \& Security Germany AG\\
       michael.brauckmann@idemia.com}
\and
\hspace{1.9cm} \IEEEauthorblockN{Tobias Glasmachers}
\IEEEauthorblockA{\hspace{1.9cm}Institute for Neural Computation\\
    \hspace{1.6cm} Ruhr-University Bochum, Germany\\
	\hspace{1.6cm}        tobias.glasmachers@ini.rub.de}
}

%


\maketitle

\begin{abstract}
This paper presents a module of vehicle re-identification based on make/model and color classification. It could be used by the Automated Vehicular Surveillance (AVS)
or by the fast analysis of video data. Many of problems, that are related to this topic, had to be addressed. In order to facilitate and accelerate the progress
in this subject, we will present our way to collect and to label a large scale data set. We used deeper neural networks in our training. They showed a good 
classification accuracy. We show the results of make/model and color classification on controlled and video data set. We demonstrate with the help of a developed application 
the re-identification of vehicles on video images based on make/model and color classification. This work was partially funded under the grant.
\end{abstract}


%
\IEEEpeerreviewmaketitle

\section{Introduction}

The objective of the vehicle re-identification module based on make/model and color classification is to recognize a vehicle within a large image or video data set based on its make/model and color attributes. There are a number of challenges related to this task, that need to be addressed:

\begin{itemize}
	\item There are more than 150 car manufacturers worldwide with approximately 2.000 models.
	\item Each model of a make generally has a longer history while model upgrades appear every few years.
	\item The appearance of a model of a make varies not only due to its model year but also differs strongly depending on the perspective. The same vehicle looks very different from the front than from the rear or from the side view.
	\item Video data frequently contains objects at low resolution, this aggravates the classification.
	\item Occlusion in case vehicles are close each to other.
\end{itemize}
In this work, we used convolutional neural networks (CNNs) to learn the vehicle's make/model and color descriptors. We used Tensorflow as framework. Our Training is applied on the detected region of interest (ROI) of the vehicle.

\section{Related Works}
Some research has been performed on make/model and/or color classification of vehicles. Most of it operates on a few number of make/models because it is difficult to get a labeled data set with all existing make/models. Manual annotation is almost impossible because one needs an expert for each make being able to recognize all its models and it is very long process. \cite{paper14} developed a make/model classification based on feature representation for rigid structure recognition using 77 different classes. They tested two distances, the dot product and the euclidean distance. \cite{paper5} tested different methods by make/model classification of 86 different classes on images with side view. The best one was HoG-RBF-SVM. \cite{paper17} use 3D-boxes of the image with its rasterized low-resolution shape and information about the 3D vehicle orientation as CNN-input to classify 126 different make/models. The module of \cite{paper10} is based on 3D object representations using linear SVM classifiers and trained on 196 classes. In a real video scene all existing make/models could occur. Considering that we have worldwide more than 2000 models, make/model classification trained on few classes will not succeed in practical applications. \cite{paper1} increases the number of the trained classes. The method is based on CNN and  trained on 59 different vehicle makes as well as on 818 different models. The solution seems to be closer to commercial use. We tried to collect the vast majority of known vehicle make/models. Our module could recognize 137 different vehicle makes as well as 1447 different models. We optimized our CNN-net architecture and got very good results on our testing data. We believe that our module gives the appropriate solution for commercial use. \cite{paper2} used color classification trained on 8 different colors. They showed an accuracy of 94.47\% in average on Stanford data set. The execution time of their module on CPU (1 core) is 3.248s. Our color classifier could recognize 10 different colors like \cite{paper1}. We used in our training images with about all existing vehicle make/models. We measured an accuracy of 97.96\% on Stanford data set. The execution time of our module for color classification on CPU (1 core, i7-4790 3.60GHz) is just 20ms. Our color classifier is faster and more accurate. The module for make/model classification on CPU (1 core, i7-4790 3.60GHz) needs also just 20ms.

\section{Training And Testing Data}
The training of Convolutional Neural Networks (CNNs) requires in general a large set of representative labeled image data. This holds true especially for vehicle classification due to the variability of the appearance of vehicles depending on the make, the model, the model year, the color and the perspective. We collected a list of the most prevalent vehicle make/models each with different views. We used a web crawler to get the vehicle images for training and testing. There is a need for data cleansing since results returned from the web contained images of the passenger compartment and other unrelated content. We made a cleansing of the data using a vehicle quality. Because of its size, this labeling of the data could not be done purely manually. To support and accelerate the labeling process, we developed a semi-automatic tool. We set a part of the collected controlled data to measure the accuracy of make/model and color classification. To test uncontrolled data, we used an internal video data set and other video data recorded by the funded research project of Victoria. Accuracy of the re-identification has been evaluated manually on video data.

\begin{figure}[bth]
	\centering
	\begin{minipage}[c]{0.10\textwidth}
		\includegraphics[width=\linewidth]{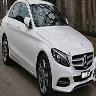}
	\end{minipage}
	\begin{minipage}[c]{0.10\textwidth}
		\includegraphics[width=\linewidth]{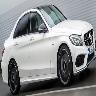}
	\end{minipage}
	\begin{minipage}[c]{0.10\textwidth}		
		\includegraphics[width=\linewidth]{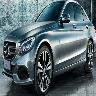}
	\end{minipage}
	\begin{minipage}[c]{0.10\textwidth}
		\includegraphics[width=\linewidth]{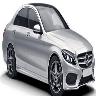}
	\end{minipage}
	\vspace*{2mm}
	\caption{Some examples of the class Mercedes-Benz C.}
\end{figure}

\section{Cost Function}
The cost function and the algorithm for the optimization have a major influence on the accuracy of the classification results and on the convergence of the net. The so-called Cross-Entropie cost function is the established choice for classification problems. A distribution density function should be emulated at the output of the network. The Softmax-function produces this:
\begin{equation}\label{key}
softmax(s_{i}) = \frac{e^{s_{i}}}{\sum_{j=1}^{N} e^{s_{j}}},  i = 1, ... , N
\end{equation}
$s_{i}$ is the similarity of the class i.\\$N$ is the number of the classes.\\
The Softmax-function has the following properties:
\begin{equation}
\sum_{i=1}^{N} f(s_{i}) = 1
\end{equation}
and
\begin{equation}
f(s_{i})\ge 0, i = 1, ... , N
\end{equation}
The cross-entropy measures the quality of a model for a probability distribution.
\begin{equation}
CE = -\frac{1}{M}\sum_{k=1}^{M}\sum_{i=1}^{N}y_{ik}\log(softmax(s_{ik}))
\end{equation}
$k$ is the index of the element in the batch.\\
$M$ is the batch size.\\\\
The label $y_{ik}$ takes a value of 1 for its corresponding class and a value of 0 for other classes. We could simplify the formula of the cross-entropy as follows:\\
\begin{equation}
CE = -\frac{1}{M}\sum_{k=1}^{M}\log(softmax(s_{k}))
\end{equation}
$s_{k}$ is the similarity of the corresponding class of the element $k$ in the batch.\\\\
Applying this cost function leads to a good separation between classes and achieves good classification accuracy. To minimize the Cross-Entropie, we used AdamOptimizer. It seems to be robust and stable and shows good convergence behaviors.

\section{CNN-Architectures}
We tested different CNN architectures. One of the two best CNNs, we had trained and optimized with several modifications, is based on the ResNet architecture. This net contains a residual layer at each block. The second net is GoogleNet like \cite{paper3}, which contains filters with different sizes, but does not contain any residual layers. In the output of each block, we have a concatenation of sub outputs. The original version of this net diverged after some thousands of iterations. We modified the inception module, which belong to GoogleNet, by adding the Z-Normalization and Elu as activation function. This leads to convergence and better results.
\begin{figure}[bth]
	\centering
	\includegraphics[width=\linewidth]{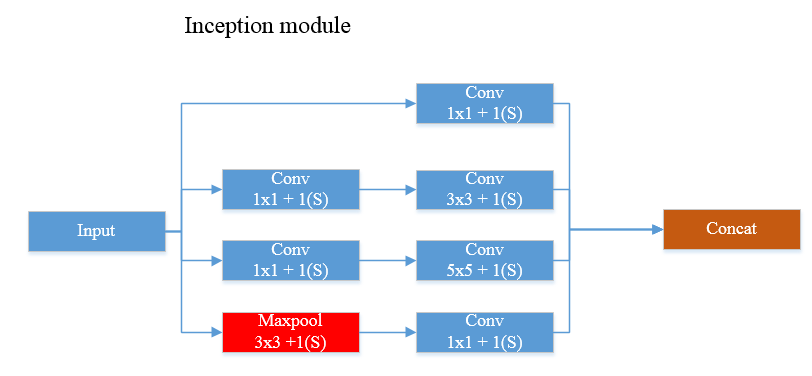}
	\caption{\label{VAE6 image} \cite{paper3} Inception module.}
\end{figure}
\begin{figure}[bth]	
	\centering
	\includegraphics[width=\linewidth]{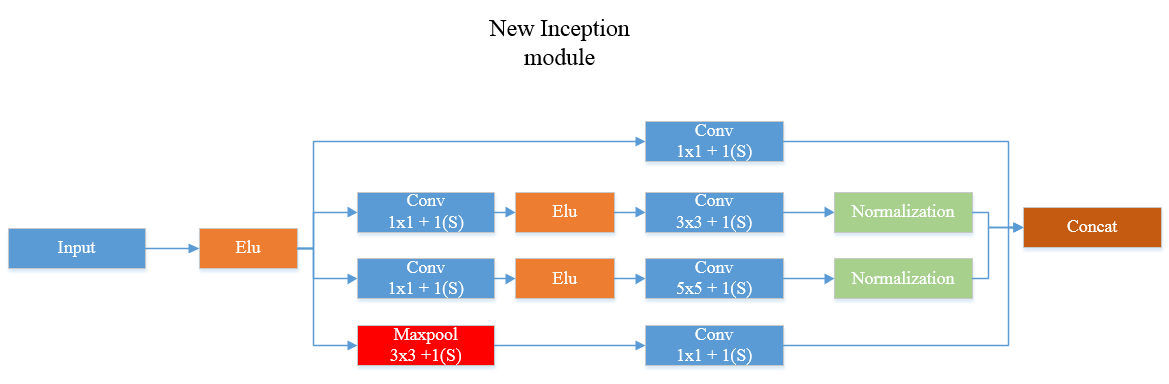}
	\caption{ Modified inception module.}
	\label{MGN}
\end{figure}

\section{Experiments}
We evaluated controlled testing data set with different views and different qualities. We used a combination of blur and down scaling to get images with bad qualities to be close to the quality of the video data. We added such images with bad quality to our training data.
\subsection{Make/Model Classification}
For make/model classification we trained 1447 classes. This allow us to classify the most make/model of vehicles worldwide. We tested both aforementioned CNN-architectures and their fusion using controlled data set with good as well as bad quality. Each data set contains 3306 images with different views and about two images per class. Our standard net is better than the modified version of GoolgeNet shown in the figure \ref{MGN}. The fusion achieves an absolute improvement of about 1.2\% for data with good quality and 2.4\% for data with bad quality. This corresponds to a residual error reduction for 8\% for data with good quality and 14.2\% for data with bad quality.
\begin{figure}[bth]
	\centering
	\includegraphics[width=\linewidth]{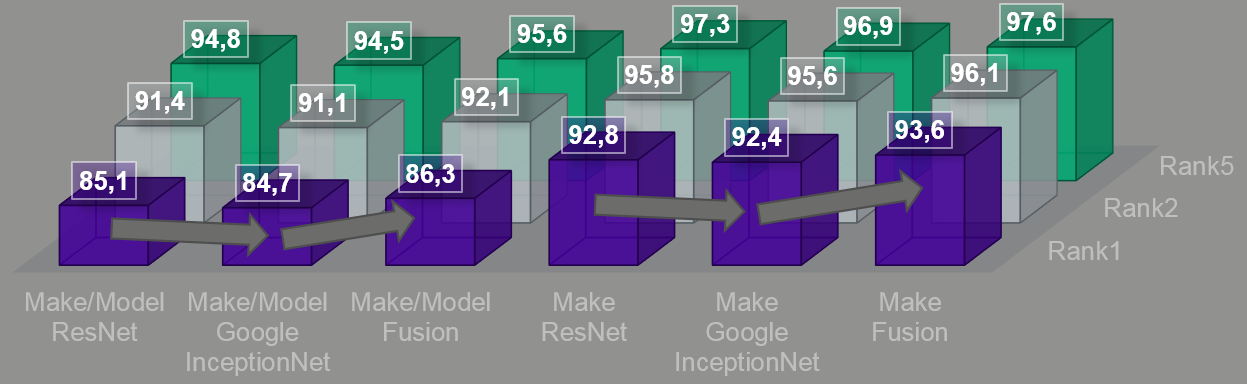}
	\caption{ Results of make/model classification on controlled data set with good quality and different views.}
\end{figure}
\begin{figure}[bth]
	\centering
	\includegraphics[width=\linewidth]{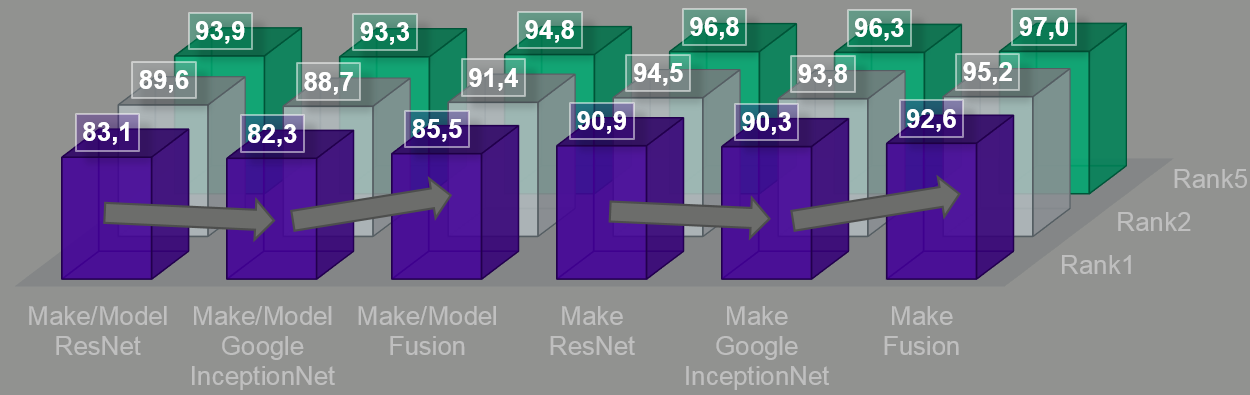}
	\caption{ Results of make/model classification on controlled data set with bad quality and different views.}
\end{figure}

\subsection{Color Classification}
\begin{figure}[bth]
	\centering
	\includegraphics[width=\linewidth]{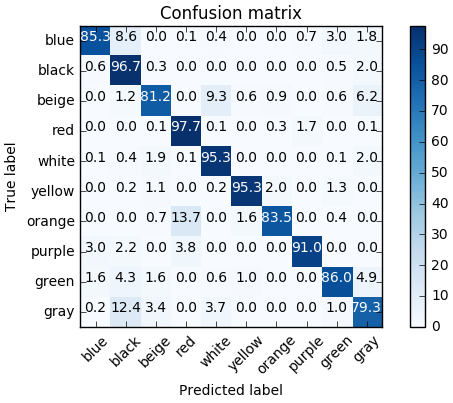}
	\caption{ The confusion matrix shows the color classification and error rates of each color on images with mixed quality (internally data set).}
	\label{conf}
\end{figure}

We optimized a CNN for color classification based on ResNet architecture. First evaluations of the accuracy of the color classification were done using two controlled data sets with good and bad quality. Each data set contains 5934 images with about all existing make/models. Both data sets contain images with different views. The figure \ref{conf} shows the correct color classification and the error rates of each class. We used ten classes like \cite{paper1} in our training. We tested also the Stanford data set and we got higher accuracy than the accuracy presented in \cite{paper2}. We did not use any part of Stanford data set in our training.


\begin{figure}[bth]
	\begin{center}
		\begin{minipage}[c]{0.22\textwidth}
			\includegraphics[width=\linewidth]{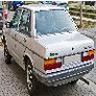}
			\caption{ Sample image with good quality.}
		\end{minipage}
		\begin{minipage}[c]{0.22\textwidth}
			\includegraphics[width=\linewidth]{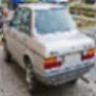}
			\caption{ Sample image with bad quality.}
		\end{minipage}
	\end{center}
\end{figure}

\begin{figure}[bth]
	\centering
	\includegraphics[width=\linewidth]{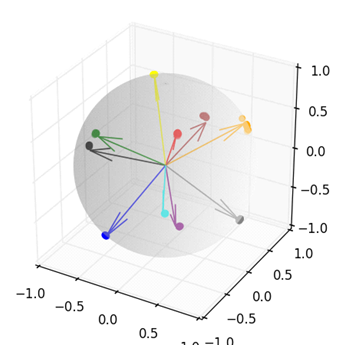}
	\caption{ The plot of the trained centroids and some feature vectors with the best classification scores on the sphere with the radius 1 (brown for white and cyan for beige).}
\end{figure}

\begin{table}[htb]
	\caption{Accuracy of our model and of the model presented by \cite{paper2} on Stanford data set.} 
	\begin{center}
		\begin{tabular}{ ?{0.8pt}c?{0.8pt}c?{0.8pt}c?{0.7pt} }
			\arrayrulecolor{colT3}\hline
			\hlineB{1.5}\cellcolor{colT1} &\cellcolor{colT1} Our color model & \cellcolor{colT1} \cite{paper2} \\ 
			\hline
			\hlineB{1.6}\cellcolor{colT1} Color correct classification & \cellcolor{colT2} 97.96\% & \cellcolor{colT2} 94.47\% \\
			\hline			
			\hlineB{1.6}\cellcolor{colT1} Execution time (CPU 1 core)& \cellcolor{colT2} 20ms & \cellcolor{colT2} 3.248s \\  
			\hlineB{1.4}\hline
			
		\end{tabular}
	\end{center}
\end{table}


\subsection{Testing of Make/Model and Color Classification on Video Data}
We processed some video data set provided in the funded research project Victoria. We applied the make/model and color classification on the best-shot images. The images contain vehicles with different views. There are problems with low resolution, lighting and in some images with occlusion. The color classification demonstrated a very robust behavior on video images. Even though we get many problems in a real scene and the classification of 1447 different classes, our make/model classification showed good results on video data. Another problem is that some models from different makes look very similar as they share the same classid like Renault/Logan and Dacia/Logan.

\begin{figure}[ht]
	\centering
	\includegraphics[width=\linewidth]{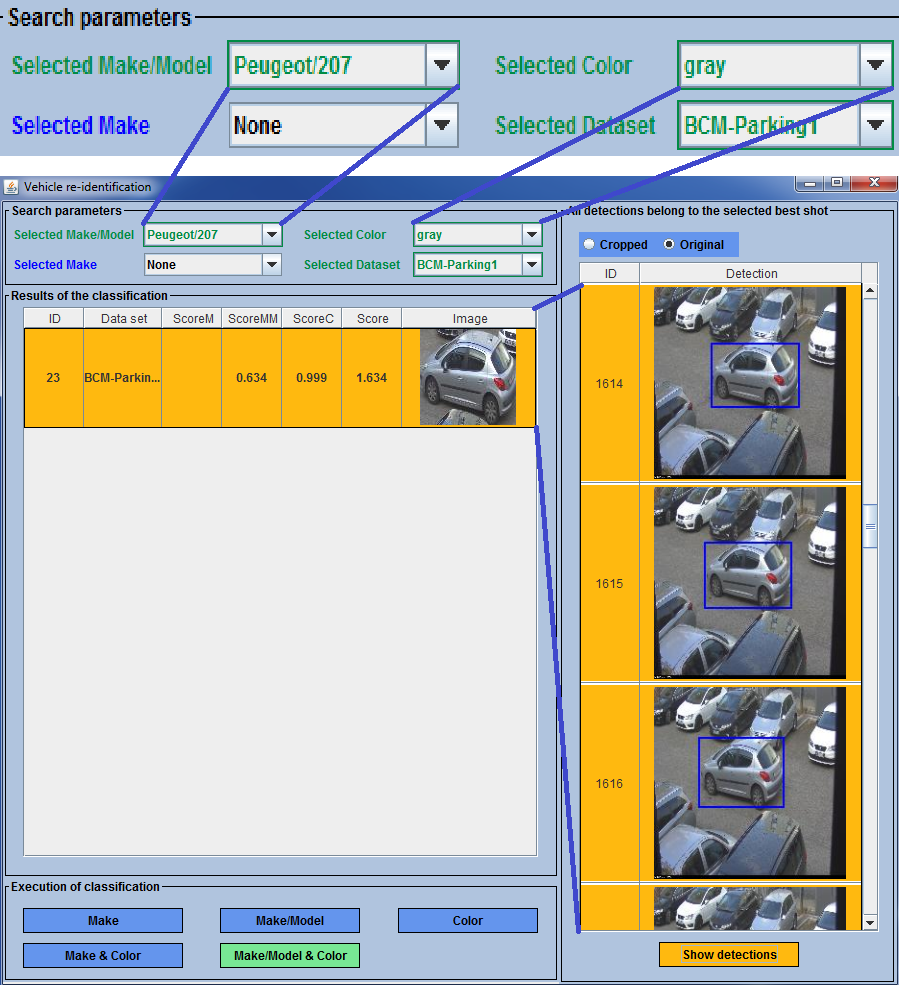}
	\caption{ Vehicle re-identification based on shape (Peugeot/207) and color (gray) classification (video data).}
\end{figure}

\begin{figure}[ht]
	\centering
	\includegraphics[width=\linewidth]{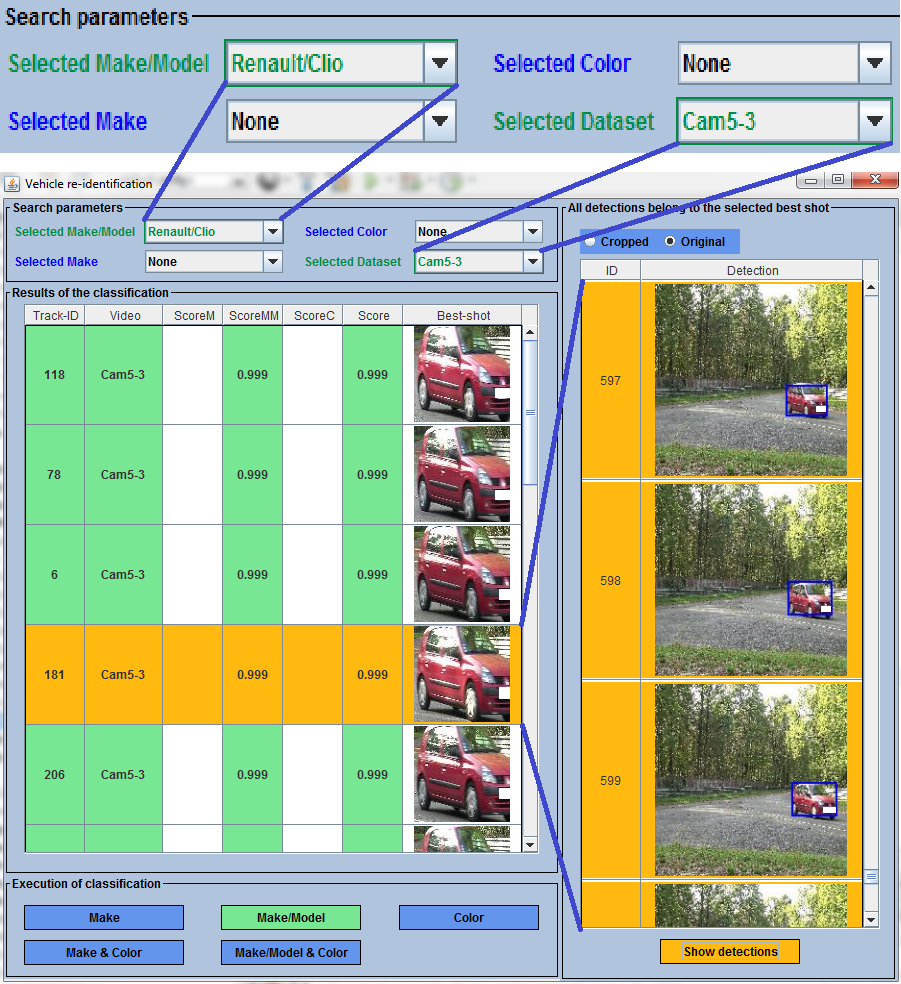}
	\caption{ Vehicle re-identification based on shape (Renault/Clio) classification (video data).}
\end{figure}

\begin{figure}[ht]
	\centering
	\includegraphics[width=\linewidth]{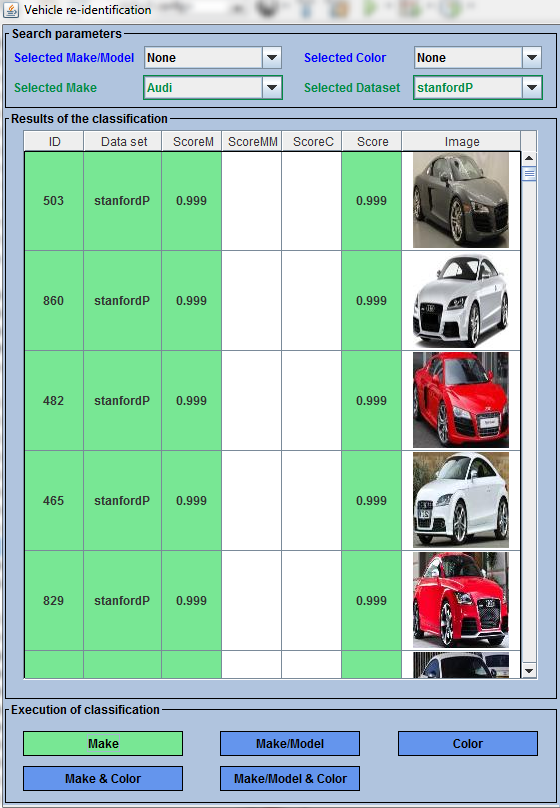}
	\caption{ Vehicle re-identification based on shape (Audi) classification (Stanford data set).}
\end{figure}

\begin{figure}[ht]
	\centering
	\includegraphics[width=\linewidth]{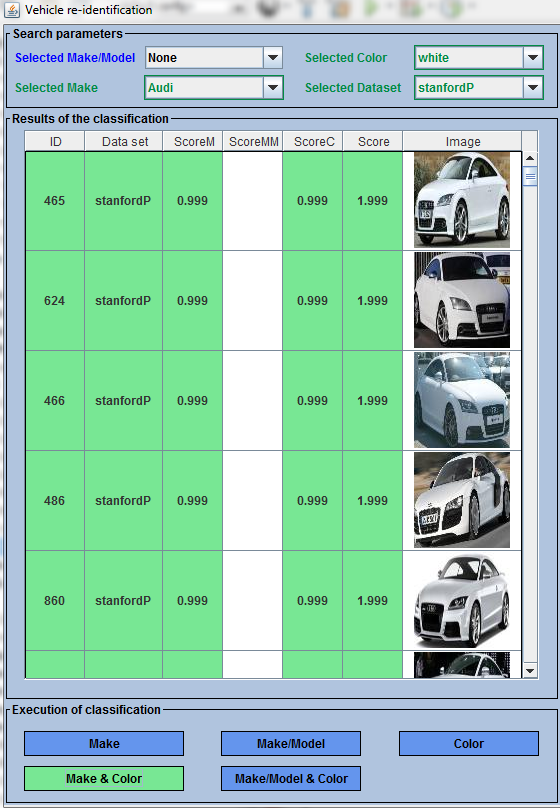}
	\caption{ Vehicle re-identification based on shape (Audi) and color (white) classification (Stanford data set).}
\end{figure}

\section{Conclusion}
The color classification is robust and shows very good results on controlled and uncontrolled data. The make/model classification shows very good results on controlled data and good results on video data. Of course, it depends on the quality of the video images and of the detections and whether a make/model of the testing image is included in the training or not. Make/Model classification needs more effort than color classification. We need to collect data of classes that are still missing in our training especially classes with older model year and we need each year to collect data of new make/models and to retrain our net.\\

\section{Acknowledgment}
\begin{itemize}
	\item Victoria: funded by the European Commission (H2020), Grant Agreement number 740754 and is for Video analysis for Investigation of Criminal and Terrorist Activities.
	\item Florida: funded by the German Ministry of Education and Research (BMBF).
\end{itemize}


\ifCLASSOPTIONcompsoc




%




	
\nocite{paper1}
\nocite{paper2}
\nocite{paper3}
\nocite{paper5}

\bibliographystyle{plain}
\bibliography{./Bibliography/biblio}
\end{document}

